\documentclass[a4paper,10pt]{article}
\usepackage[utf8]{inputenc}
\usepackage{amsmath, amssymb, amsthm}
\usepackage{amsmath}
\usepackage{graphicx}
\usepackage[style=authoryear, sorting=nyt, maxcitenames=1]{biblatex}
\usepackage{multicol}
\usepackage{enumitem} 
\usepackage{wrapfig}
\usepackage{multirow}
\usepackage{float}
\usepackage{booktabs}
\usepackage{subcaption}
\usepackage{makecell}

\title{Class-specific Data Augmentation for Plant Stress Classification}

\author
{Nasla Saleem$^{1}$, Aditya Balu$^{1}$, Talukder Zaki Jubery$^{1}$, Arti Singh$^{2}$, Asheesh K. Singh$^{2}$, \\Soumik Sarkar$^{1}$, Baskar Ganapathysubramanian,$^{1\ast}$ 
\\
\normalsize{$^{1}$Department of Mechanical Engineering, Iowa State University, Ames, IA, USA}\\
\normalsize{$^{1}$Department of Agronomy, Iowa State University, Ames, IA, USA}\\
\normalsize{$^\ast$To whom correspondence should be addressed; E-mail:  baskarg@iastate.edu}
\\
}

\begin{document}
\maketitle

\textbf{Core Ideas}

\begin{itemize}
\item We propose an effective approach for automated selection of class-specific data augmentations for precise plant stress classification. 

\item  Employing a genetic algorithm for efficient augmentation strategy selection in challenging datasets.

\item Achieving significant performance gains with reduced computation via finetuning only the linear layer of the CNN-model

\end{itemize}


\begin{abstract}
Data augmentation is a powerful tool for improving deep learning-based image classifiers for plant stress identification and classification. However, selecting an effective set of augmentations from a large pool of candidates remains a key challenge, particularly in imbalanced and confounding datasets.  We propose an approach for automated class-specific data augmentation using a genetic algorithm. We demonstrate the utility of our approach on soybean [\textit{Glycine max (L.) Merr}] stress classification where symptoms are observed on leaves; a particularly challenging problem due to confounding classes in the dataset. Our approach yields substantial performance, achieving a mean-per-class accuracy of 97.61\% and an overall accuracy of 98\% on the soybean leaf stress dataset. Our method significantly improves the accuracy of the most challenging classes, with notable enhancements from 83.01\% to 88.89\% and from 85.71\% to 94.05\%, respectively.

A key observation we make in this study is that high-performing augmentation strategies can be identified in a computationally efficient manner. We fine-tune only the linear layer of the baseline model with different augmentations, thereby reducing the computational burden associated with training classifiers from scratch for each augmentation policy while achieving exceptional performance. This research represents an advancement in automated data augmentation strategies for plant stress classification, particularly in the context of confounding datasets. Our findings contribute to the growing body of research in tailored augmentation techniques and their potential impact on disease management strategies, crop yields, and global food security. The proposed approach holds the potential to enhance the accuracy and efficiency of deep learning-based tools for managing plant stresses in agriculture.
\end{abstract}



\section{Introduction}
    
    Accurate classification of plant stresses is of utmost importance for effective crop management and sustainable agricultural practices \parencite{al2011fast}. Both biotic (diseases and insects) and abiotic plant stresses (drought, salinity, temperature extremes, and nutrient deficiencies) have detrimental effects on crop growth, yield, and quality \parencite{mosa2017introduction}. By precisely identifying and classifying these stresses early, farmers can develop targeted strategies to mitigate their impact and optimize crop health \parencite{sankaran2010review,nagasubramanian2018hyperspectral}. Moreover, accurate stress classification plays a key role in selecting stress-tolerant crop varieties \parencite{singh2021plant} and can make a significant impact on improved genomic studies and high-throughput phenotyping \parencite{SINGH2016110, singh2018deep, zhang2017computer}.  Accurate stress classification can enhance cyber-agricultural systems, leading to improved crop resilience, reduced production losses, and sustainable agricultural practices\parencite{gao2020deep, gill2022comprehensive, gonzalez2022new}. In this paper, we explore the development of accurate classifiers for plant stress classification, aiming to improve downstream plant stress management activities involving stress identification and enable effective mitigation strategies.
    
    \begin{figure}[ht]
       \includegraphics[width=\textwidth, trim=10 260 10 250, clip]{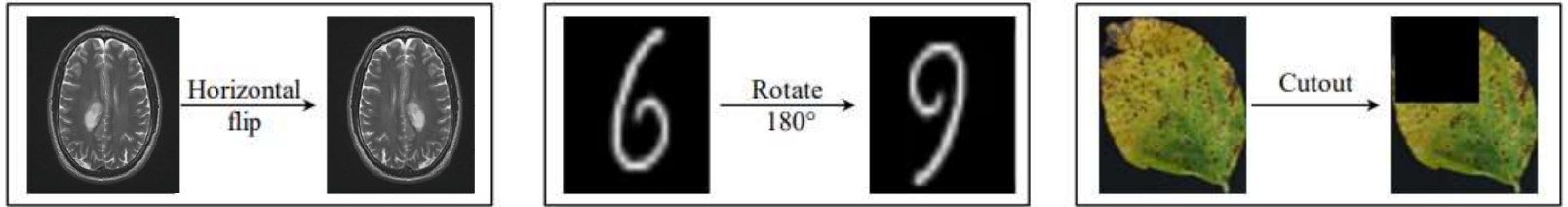}
        \caption{Class-specific effects of augmentations: \textit{"horizontal flip"} distorts a brain cell image, \textit{"vertical flip"} transforms a "6" into a "9" in MNIST, and \textit{"cutout"} masks disease in a soybean leaf. These instances reveal that tailored strategies are essential, as not all augmentations benefit all classes.
            \label{fig: class-specific effects}}
    \end{figure}
    Traditionally, plant stress identification and quantification heavily relied on the expertise of human scouts and domain experts \parencite{SINGH2016110}. However, this manual approach is time-consuming, subjective, and limited in scalability, posing challenges in terms of efficiency and accuracy. The emergence of advanced technologies such as drones \parencite{xu2023instance, feng2021comprehensive, guo2021uas, herr2023unoccupied}, ground robots \parencite{atefi2021robotic, gao2018novel}, and sensors \parencite{parmley2019development, pieruschka2019plant}, has brought high-throughput phenotyping and phenomics to the forefront \parencite{araus2014field}, transforming the measurement of multiple plant traits across various growth stages and facilitating rapid, precise, and accurate data collection. Machine learning (ML) and deep learning (DL) techniques have emerged as effective tools in automating plant stress classification processes \parencite{ghosal2018explainable, SINGH2016110}. Despite promising outcomes in discerning various plant stresses, DL models encounter a significant challenge: the requirement for abundant labeled and diverse data \parencite{kamilaris2018deep}. To address this challenge, data augmentation (DA) has emerged as a valuable approach to enhancing model performance by augmenting the available data through various transformations \parencite{van2001art, krizhevsky2012imagenet, shorten2019survey}. These transformations include rotation, flipping, scaling, cropping, and noise injection, effectively minimizing performance gaps between training and testing stages, reducing overfitting, and improving the generalization capability of DL models \parencite{shorten2019survey, rebuffi2021data, taylor2018improving}. Importantly, data augmentation allows for effectively expanding the training data without the need for laborious manual labeling or extensive data collection efforts, making DL models more accessible and efficient for plant stress classification tasks \parencite{8628742}.
    
    \begin{figure}[ht]
       \includegraphics[width=0.9\textwidth, trim=65 30 65 25, clip]{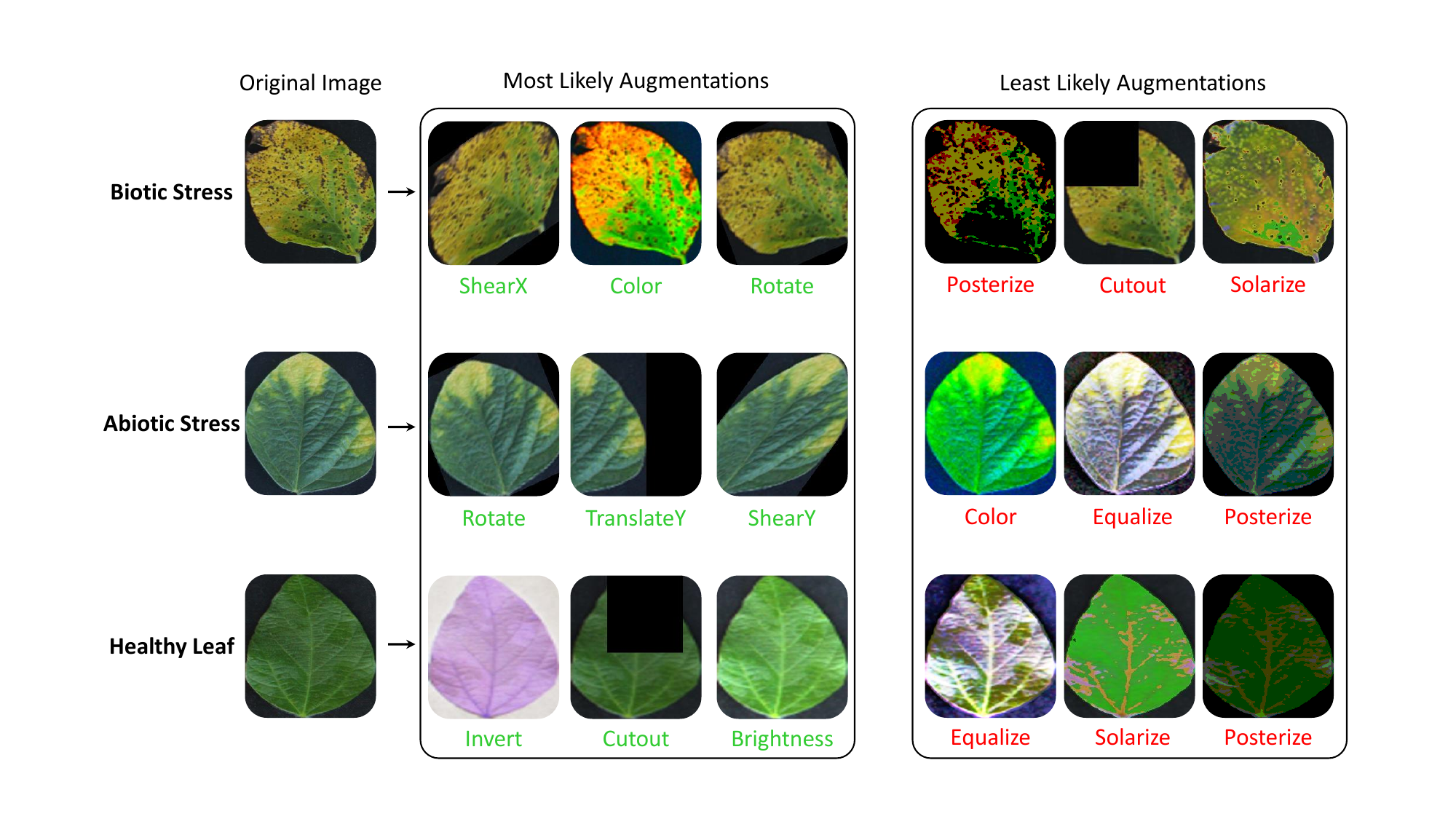}
        \caption{For different stress classes in the soybean stress (biotic and abiotic) dataset, we present an image from each category (left) and thin automating plant stress classification processes e corresponding image transformed using the three most likely augmentations (middle) and the three least likely augmentations (right) for that stress class, as determined by our class-specific automated data augmentation method.
            \label{fig: Augs chosen by DA}}
    \end{figure}
    
    Despite the effectiveness of data augmentation in enhancing the performance of DL models, manually selecting appropriate augmentation techniques is time-consuming and challenging. To address this issue, researchers have turned to automated machine learning (AutoML) \parencite{he2021automl} techniques for automatically searching and selecting augmentation policies on datasets \parencite{cubuk2018autoaugment, ho2019population, lim2019fast, zoph2020learning}. These include methods like AutoAugment \parencite{cubuk2018autoaugment}, Fast AutoAugment \parencite{lim2019fast}, and Faster AutoAugment  \parencite{weng2019gan}, which use reinforcement learning or density matching to find optimal augmentation policies \parencite{terrell1992variable}. Additionally, gradient-based methods such as DeepAutoAugment \parencite{zheng2022deep} automate policy selection without prior knowledge. These methods directly learn the augmentation policy without prior knowledge or manual selection of default transformations such as \parencite{marrie2023slack}.  However, the computational complexity of these methods limits their feasibility for image classification problems with limited computational resources and time constraints. Population-based augmentation (PBA) is another promising technique that enables the simultaneous training and evaluation of multiple augmentation policies, facilitating efficient policy discovery \parencite{ho2019population}.  Notably, PBA has demonstrated effectiveness in discovering diverse and high-performing augmentation policies while imposing minimal computational overhead. In our study, we specifically opted for PBA due to its superior efficiency and effectiveness and further explored its potential for augmentation policy selection on a class-specific basis. It's important to note that while these methods search for policies suitable for the entire dataset, the class-dependent nature of augmentation policies has received limited attention in current research. Although the generation of class-dependent data has been studied in the context of GANs \parencite{mirza2014conditional}, to our knowledge, only a few works have explored class-dependent data augmentation \parencite{hauberg2016dreaming, rommel2021cadda}.
    
    While data augmentation is commonly employed to enhance model performance, different classes within a dataset may exhibit varying sensitivities to specific transformations \parencite{balestriero2022effects}. This discrepancy in sensitivity becomes especially pronounced in scenarios where certain classes are subjected to per-class favoritism, leading to biased predictions and arbitrary inaccuracies on specific classes. For example, in the context of object recognition in images, using color transformations can benefit the model's ability to recognize objects such as cars or lamps, but this same augmentation strategy may have a detrimental effect on classes that are strongly defined by their color, such as apples or oranges. Similarly, applying "vertical flip" augmentation in the MNIST dataset \parencite{deng2012mnist} alters the visual representation of classes 6 and 9 as illustrated with some other examples in Figure~\ref{fig: class-specific effects}.  
    
    This observation extends to plant stress classification, where distinguishing between different stress types and healthy plants can be challenging due to subtle visual differences. Several works have aimed to enhance detection accuracy using practical data augmentation technique \parencite{cap2020leafgan,zhu2020data,pawara2017data}. For instance, in cases of potassium deficiency, early identification is crucial as leaf yellowing starts from the tip of soybean leaflets. However, using cutout \parencite{devries2017improved} augmentation targeting the tip of the potassium-stressed leaf might compromise model's ability to identify potassium deficiency early on. Additionally, datasets with confounding classes (classes that are difficult to distinguish from one another due to overlapping visual characteristics or shared features)  pose an additional challenge, as data augmentation can potentially worsen performance disparities among classes. Thus, applying transformations that emphasize texture or shape features to classes that are difficult to distinguish can be beneficial. Consequently, class-specific data augmentation emerges as a potent tool for enhancing ML model performance, particularly in scenarios with challenging classes.  
    
    To address the challenges posed by class-dependent invariances and to enhance classifier performance, particularly for confounding classes, we propose a novel approach that customizes augmentation strategies to capture the unique characteristics of each class. By fine-tuning a pre-trained image classification model and optimizing augmentation policies for individual classes, our class-specific approach aims to improve mean-per-class accuracy, particularly in the context of confounding classes in the dataset. The automated process of class-specific data augmentation, driven by an evolutionary optimization algorithm,  Genetic Algorithm (GA),  \parencite{katoch2021review}, selects the most effective augmentation policies for each class. 

    The effectiveness of our approach is demonstrated in Figure~\ref{fig: Augs chosen by DA}, where transformed images using the most and least likely augmentations for each stress class are visualized. These results highlight the efficacy of class-specific data augmentation in improving model performance. This tailored strategy strikes a balance between efficiency and effectiveness, providing a promising solution to address limitations of conventional augmentation techniques.
    

    Specific contributions of this paper are summarized below as follows:
    
    \begin{itemize}
    \item We propose an effective approach based on GA to find the best set of augmentations for each class on a target dataset. 
    
    \item We demonstrate the efficacy of our approach by showing that our per-class augmentations significantly improved the accuracy of the two worst-performing classes in the target dataset, increasing from 83.01\% to 88.89\% and 85.71\% to 94.05\%. Additionally, our approach significantly increased the mean-per-class accuracy of the dataset from 95.09\% to 97.61\% compared to the accuracy of the non-augmented model.
    \end{itemize}
    
    In our implementation, a well-trained classifier is used as a baseline, and it is fine-tuned for only 5 epochs with various sets of augmentations whose probabilities are the population created by GA. This approach significantly reduces the computational cost of searching for optimal augmentation policies while maintaining competitive performance.

\section{Materials and Methods}

    \subsection{Dataset}
    
\begin{figure}[ht]
           \includegraphics[width=0.9\textwidth, trim=130 75 150 70, clip]{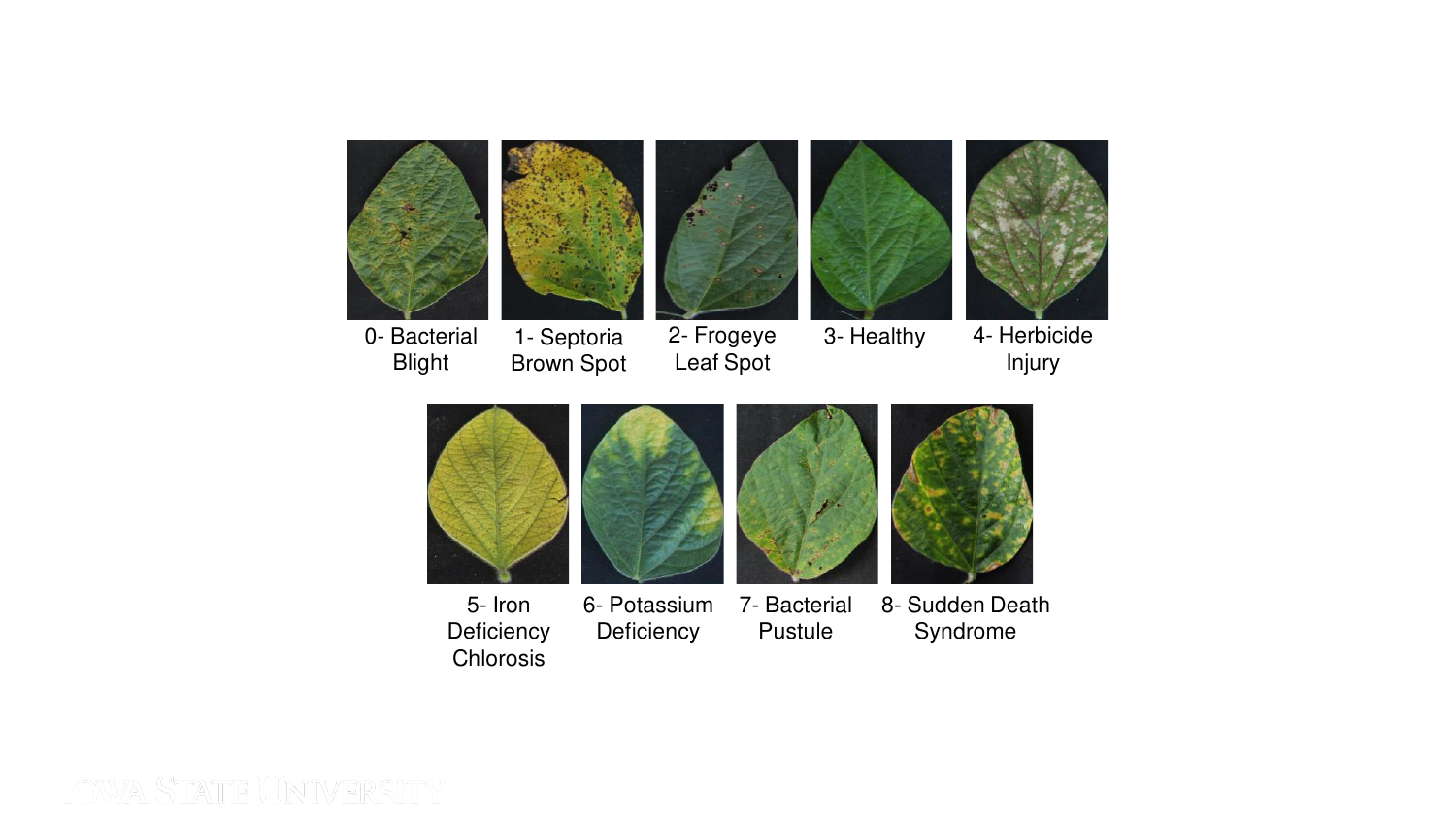}
            \caption{Image examples of the nine classes (healthy leaflet and eight different soybean stresses) in the dataset.
                \label{fig: Image examples}}
        \end{figure}

        The dataset used in this study is a publicly available dataset comprising 16,573 RGB images of soybean leaflets across nine distinct classes, eight different soybean stresses and healthy soybean leaflets, covering a broad range of biotic and abiotic foliar stresses \parencite{ghosal2018explainable}. Figure \ref{fig: Image examples} demonstrates the imaging setup and the nine soybean leaf stress classes included in the analysis. The training, validation, and test datasets were composed of $13,420$ (80\%), $1,491$ (9\%), and $1,662$ (11\%), respectively (Table S1). More information on the dataset is available in \parencite{ghosal2018explainable}.
        
    \subsection{Baseline model}
    
        To establish a fair and comprehensive baseline for soybean stress classification, we followed the methodology outlined in \parencite{ghosal2018explainable}, utilizing the same dataset and model. However, to explore potential improvements, we experimented with different DL architectures to enhance the baseline accuracy. Our findings, detailed in Table S2, revealed that ResNet50 \parencite{he2016deep} achieved the highest accuracy (95.09\%) surpassing the previously reported 94.13\%, prompting its selection for further evaluation. The baseline model was trained for 350 epochs on the dataset without any data augmentations to ensure unbiased performance \parencite{balestriero2022effects}. During training, we utilized the categorical cross-entropy loss function, Adam optimizer with a momentum of 0.9, weight decay of 0.0001, and a batch size of 256.
        
        Despite the promising results obtained with the baseline model, a consistent observation akin to the study's findings emerged, wherein the model encountered difficulties in accurately classifying the challenging categories of Bacterial Blight and Bacterial Pustule with per-class accuracies of 83.01\% and 85.71\%, respectively. Discriminating  between  these two stresses is challenging even for expert plant pathologists due to confounding symptoms \parencite{hartman2015compendium}.

\begin{wrapfigure}{r}{0.35\textwidth}
            \vspace{-25pt} 
            \includegraphics[width=0.4\textwidth, trim=280 175 280 150, clip]{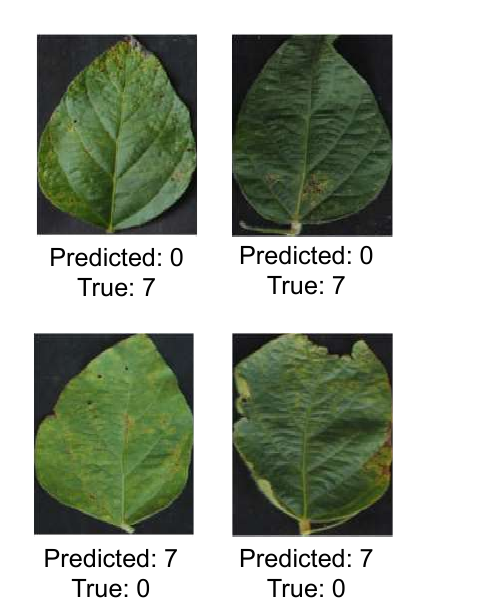}
            \caption{Examples of wrongly classified images by baseline model}
            \vspace{-10pt} 
            \label{fig:misclassified_imags}
        \end{wrapfigure}

        A few examples of misclassified images by the baseline model are provided in Figure~\ref{fig:misclassified_imags}, and from the figure, it is evident that these two stresses are hard to classify even for human experts. These findings highlight the need for further refinement and optimization, as an ideal classifier should excel in accurately predicting all classes, including those that pose significant challenges. To address this, our primary focus was to enhance the accuracy of the worst-performing classes, particularly targeting Bacterial Blight and Bacterial Pustule. The proposed GA-optimized automated DA algorithm is evaluated using our enhanced baseline model as a foundation.

    \subsection{ Genetic algorithm for optimizing data augmentations}
    
        We utilized GA, a search algorithm inspired by natural selection and genetic inheritance, to drive the evolutionary process in our study \parencite{katoch2021review}. It is a method used to find the best solution to an optimization problem by exploring a population of potential solutions. Each individual in the population represents a potential solution to the problem. Through successive generations, GA iteratively explores and evolves the population, aiming to converge toward the optimal or near-optimal solution. The effectiveness of GA in achieving this goal relies on the incorporation of elitism. Elitism ensures that the best individuals from the current generation are preserved and directly transferred to the next generation without alteration. This strategy helps maintain diversity within the population while safeguarding promising solutions from premature elimination due to the randomness of genetic operations such as mutation and crossover.
        
        
        
        
        \begin{figure}[t]
           \includegraphics[width=1\textwidth, trim=10 65 12 20, clip]{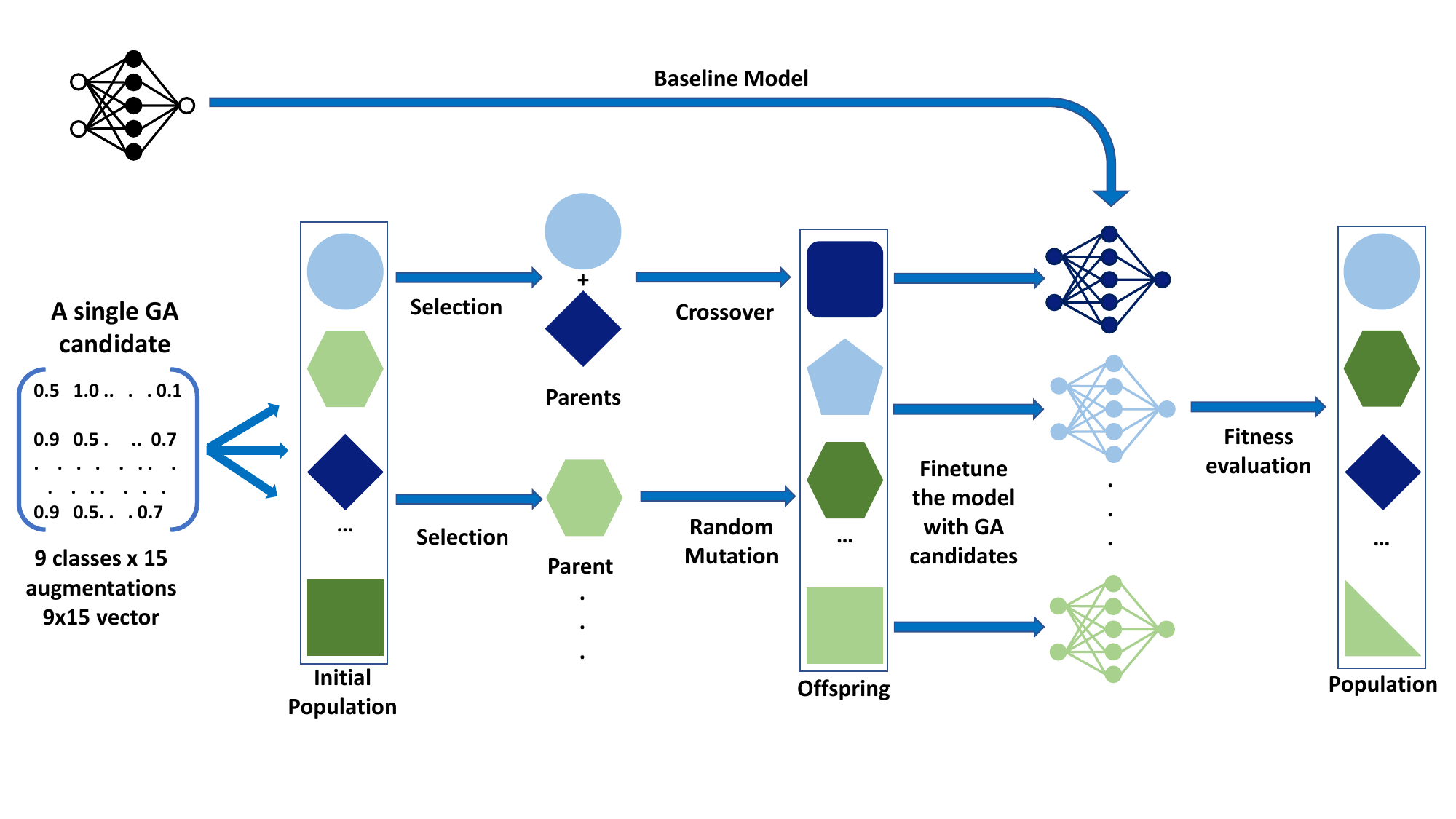}
            \caption{Illustration of a single generation in the GA framework. The baseline classifier is fine-tuned with each candidate from the GA population, which represents the probabilities of augmentations for each class. These selected candidates undergo mutation and crossover operations, generating the next generation of augmentation probabilities for improved performance.
                \label{fig:workflow}}
        \end{figure}
        
        In the context of our soybean leaf stress dataset, we employ GA to optimize the probability of each augmentation for the 9 classes. Our data augmentation search space is composed of the standard pool of 15 transformations; ShearX/Y, Translate X/Y, Rotate, AutoContrast, Invert, Equalize, Solarize, Posterize, Contrast, Color, Brightness, Sharpness, and Cutout. These augmentations, closely align with those utilized in AutoAugment \parencite{Adam_2014}, have emerged as popular choices for exploring optimal data augmentation policies in image classification tasks.

        The search space for this optimization problem consists of all possible combinations of augmentation for each of the 9 classes. To streamline our optimization process, we consider probabilities ranging from 0 to 1 with a step size of 0.1 for applying each augmentation to the respective class. By defining the augmentation magnitude as the mean of the possible values, we maintain consistency in the augmentation's influence. Our primary objective is to determine the most effective combination of augmentation probabilities for each class that maximizes the mean per-class accuracy of our target dataset.
        
        Our GA operators include:
        \begin{itemize}
            \item \textbf{Initialization:} Create an initial population set of probabilities ranging from 0 to 1 for each augmentation strategy.
            
            \item \textbf{Evaluation:} Assess the fitness of each augmentation strategy by evaluating its mean-per-class accuracy on the test dataset.
            
            \item \textbf{Selection:} Choose augmentation strategies with higher accuracy as parents for the next generation, using fitness proportionate selection or other selection strategies.
            
            \item \textbf{Crossover:} Combine probabilities of two augmentation sets to create offspring individuals with a mix of their characteristics.
            
            \item \textbf{Mutation:} Introduce random changes or modifications to the probabilities of augmentations to maintain diversity and explore new regions of the search space.
        \end{itemize}

        Formally, let $\mathbf{p} = (p_{ij})$ be a $9 \times 15$ matrix, where $p_{ij}$ represents the probability of applying the $j$-th augmentation technique to samples from the $i$-th class during training. The optimization problem can be defined as follows:
        \begin{align*}
        \text{Maximize:} & \quad \text{{MPCA}} \\
        \text{Subject to:} & \\
        \text{Constraint:} & \quad 0 \leq p_{ij} \leq 1, \quad \forall i, j \\
        \end{align*}
        The objective is to maximize the mean -per-class accuracy (MPCA) and the constraints ensure that the augmentation probabilities remain within the feasible range for each decision variable. The illustraton of our GA framework for a single generation is shown in Figure \ref{fig:workflow}. By employing GA, we aim to effectively explore and navigate this search space, searching for the set of augmentation probabilities that leads to the highest classification accuracy on our dataset. To evaluate the performance of the classifiers, we employ commonly used evaluation metrics, including overall accuracy, mean-per-class accuracy, and confusion matrix analysis.
        
        \[
        \text{{Mean-per-class accuracy (MPCA)}} = \frac{1}{N} \sum_{i=1}^{N} \text{{Accuracy}}_i
        \]
    
    \subsection{Fine-tuning baseline model with augmentation probabilities}
    
        \begin{figure}
          \centering
          \includegraphics[width=0.9\textwidth, trim=50 55 50 50, clip]{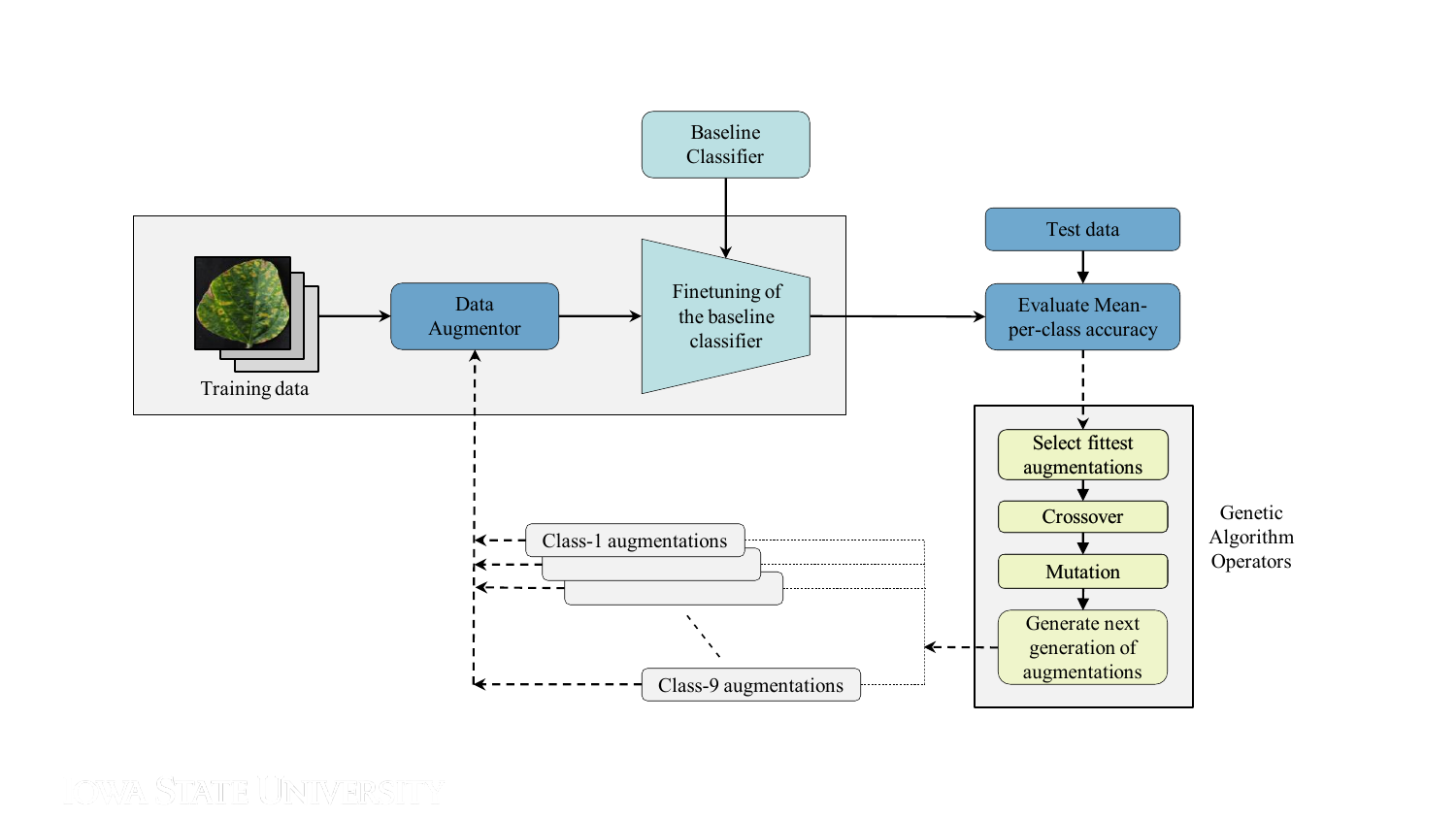}
          \caption{Flowchart depicting the overall workflow for optimizing augmentation probabilities using genetic algorithm}
          \label{fig:flowchart}
        \end{figure}
        
        Figure~\ref{fig:flowchart} illustrates the flowchart of the overall workflow for optimizing augmentation probabilities using GA. After generating a population of augmentation probabilities, the baseline model is fine-tuned for each augmentation probability in the population. The fine-tuning process involves applying the augmentation probabilities to the training data, evaluating the resulting classifier on the test set, and using the mean-per-class accuracy as the fitness score for each chromosome. Based on these fitness scores, GA performs selection, crossover, and mutation operations to generate a new population of chromosomes. This process continues iteratively until a termination criterion is met or the best solution is obtained. It is worth noting that the child networks in this study undergo a concise fine-tuning process of only 5 epochs, which is significantly shorter compared to other automated data augmentation strategies. We selected 5 epochs based on the observation of limited performance improvement beyond this point.

    \subsection{Implementation details}
    
        The experiments were conducted on a GPU cluster at Iowa State University, featuring four A100 NVIDIA GPUs, each equipped with 80 GB of memory. This configuration allowed us to concurrently fine-tune eight models by utilizing two models on each GPU, significantly reducing the overall processing time by running GA in parallel across 8 GPUs. On average, one generation took approximately 4.5 hours to complete. As a future work, further optimization can be achieved by distributing the workload across multiple nodes, which would result in even faster processing times. Our proposed method requires less computation than traditional automated DA methods since we only fine-tune the base model using a set of augmentation probabilities for five epochs. 
        
        To implement the GA, we employed PyGAD \parencite{gad2021pygad} and configured with a maximum of 100 generations. Termination criteria were defined as either completing 100 generations or observing no improvement in fitness scores for ten consecutive generations. Hyperparameters were optimized using the Rastrigin function, known for its challenging landscape characterized by multimodality and high oscillation \parencite{pohlheim2007examples}. A population size of 100 individuals was chosen for the GA, employing steady-state selection, random mutation, and single-point crossover to maintain diversity and explore the search space effectively.

\section{Results}

    The primary goal of our experiments is to assess the effectiveness of automated class-specific data augmentation using a GA-based approach in improving mean-per-class accuracy and the accuracy of worst-performing classes. We demonstrate this by evaluating the performance of the models across each class, examining the corresponding confusion matrices, analyzing augmentations selected by GA, and the impact of the order of augmentations in classification accuracy. To ensure the robustness and generalization of our model, we conducted 5-fold cross-validation on our dataset. The outcomes of this cross-validation, presented in Table S3, guided our selection of the most effective model for further investigation.
    
    \begin{figure}[t]
        \centering
        \includegraphics[width=0.9\textwidth, trim=8 7 5 7, clip]{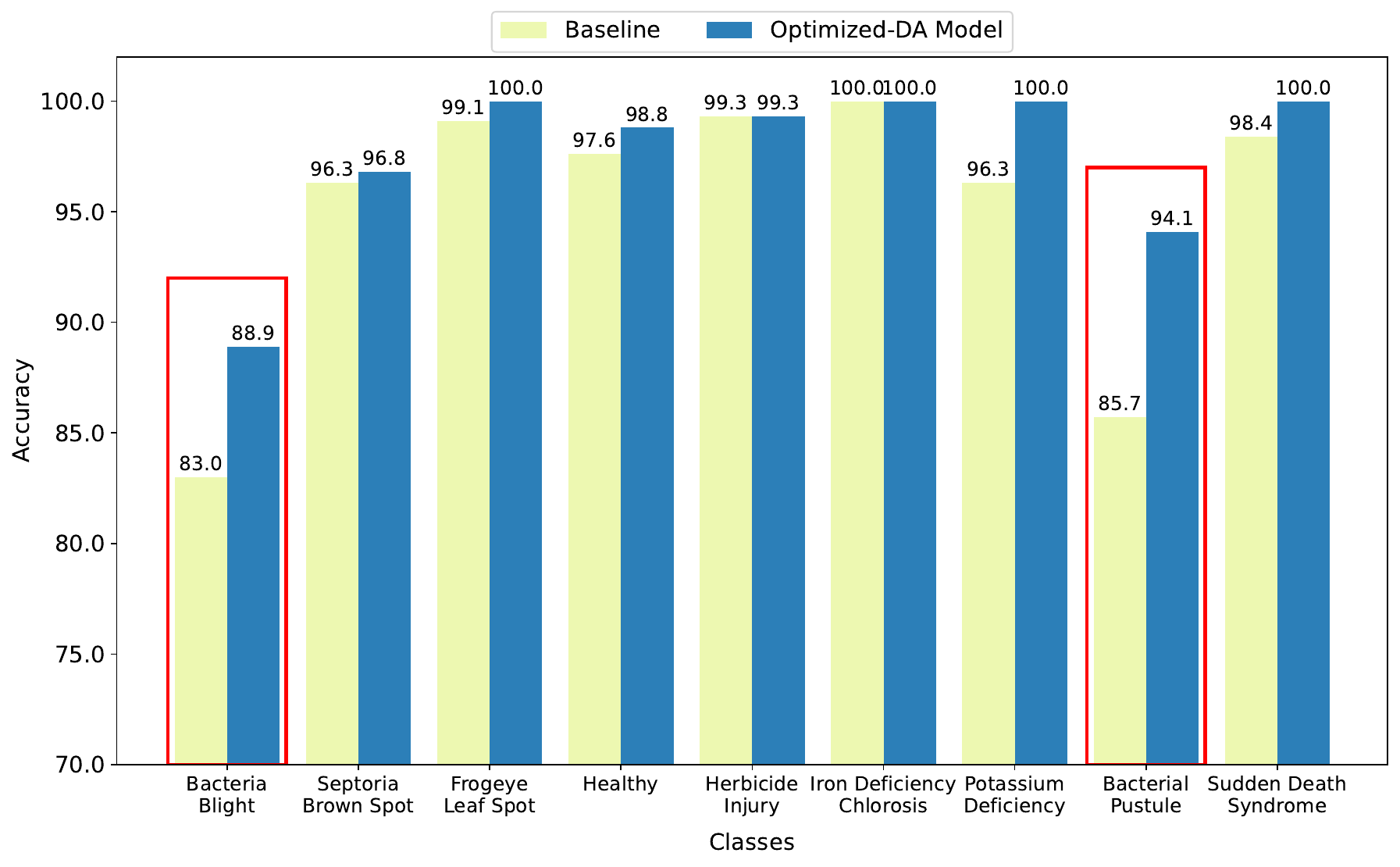}
        \caption{Comparison of class-wise accuracies: Bar chart comparing the accuracies of the baseline model and the optimized-DA model achieved through a GA-based evolutionary process. The optimized-DA model showcases remarkable improvements in accuracies for all classes, with particularly notable enhancements observed for the confounding classes - bacterial blight and bacterial pustule. 
            \label{fig: Accuracy bar chart}}
    \end{figure}
    
    \subsection{Impact of class-specific augmentations on classification accuracy}

        The bar chart in Figure~\ref{fig: Accuracy bar chart} provides a clear comparison between the baseline model and the optimized model after applying GA-based automated data augmentation. It demonstrates a substantial improvement in the mean-per-class accuracy, from 95.09\% with the baseline model to an impressive 97.61\% with the optimized model. This enhancement across all classes indicates the efficacy of employing tailored class-specific augmentations, enabling the model to better recognize and differentiate between different class characteristics, ultimately leading to more accurate classification.
        
        \begin{figure}[t]
            \centering
            \begin{subfigure}{0.49\linewidth}
                \includegraphics[width=\textwidth, trim=0 0 0 2, clip]{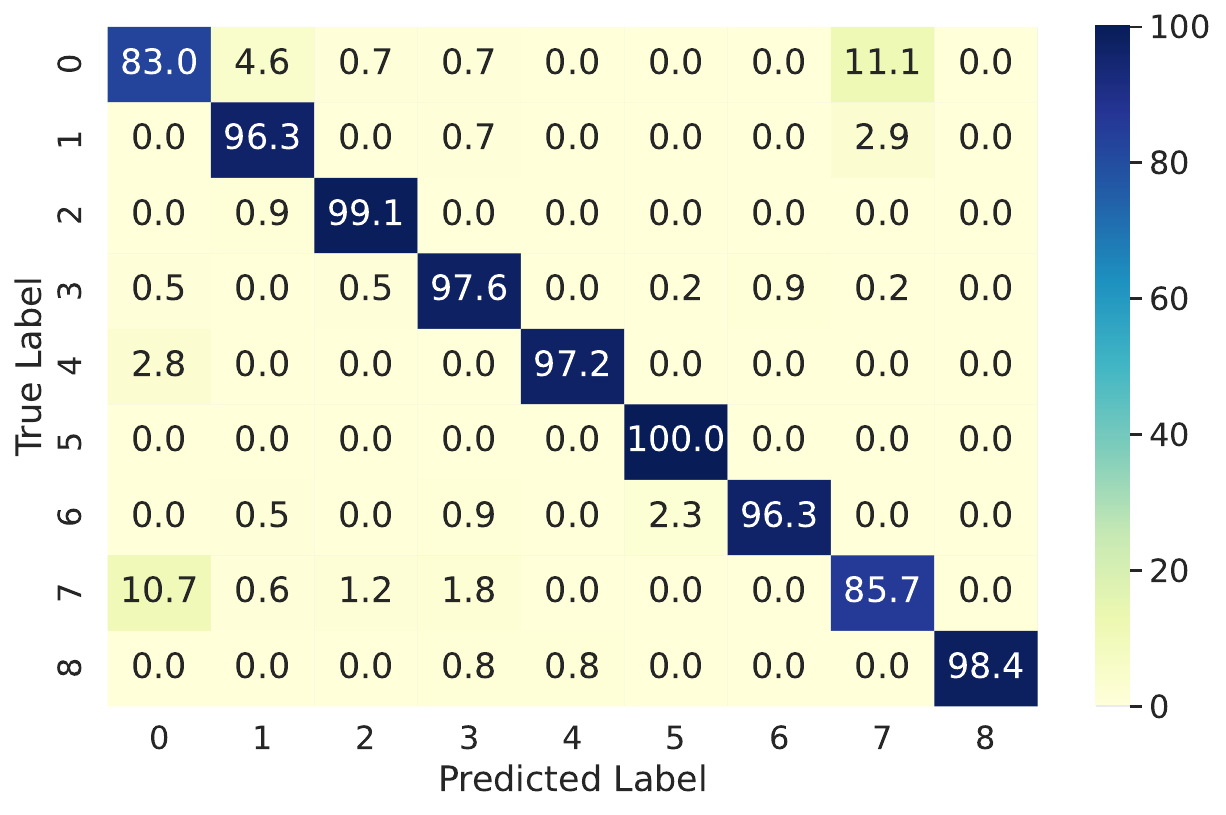}
                \caption{Baseline Model}
                \label{fig:baseline-confusion-matrix}
            \end{subfigure}
            \hfill
            \begin{subfigure}{0.49\linewidth}
                \includegraphics[width=\textwidth, trim=0 0 0 2, clip]{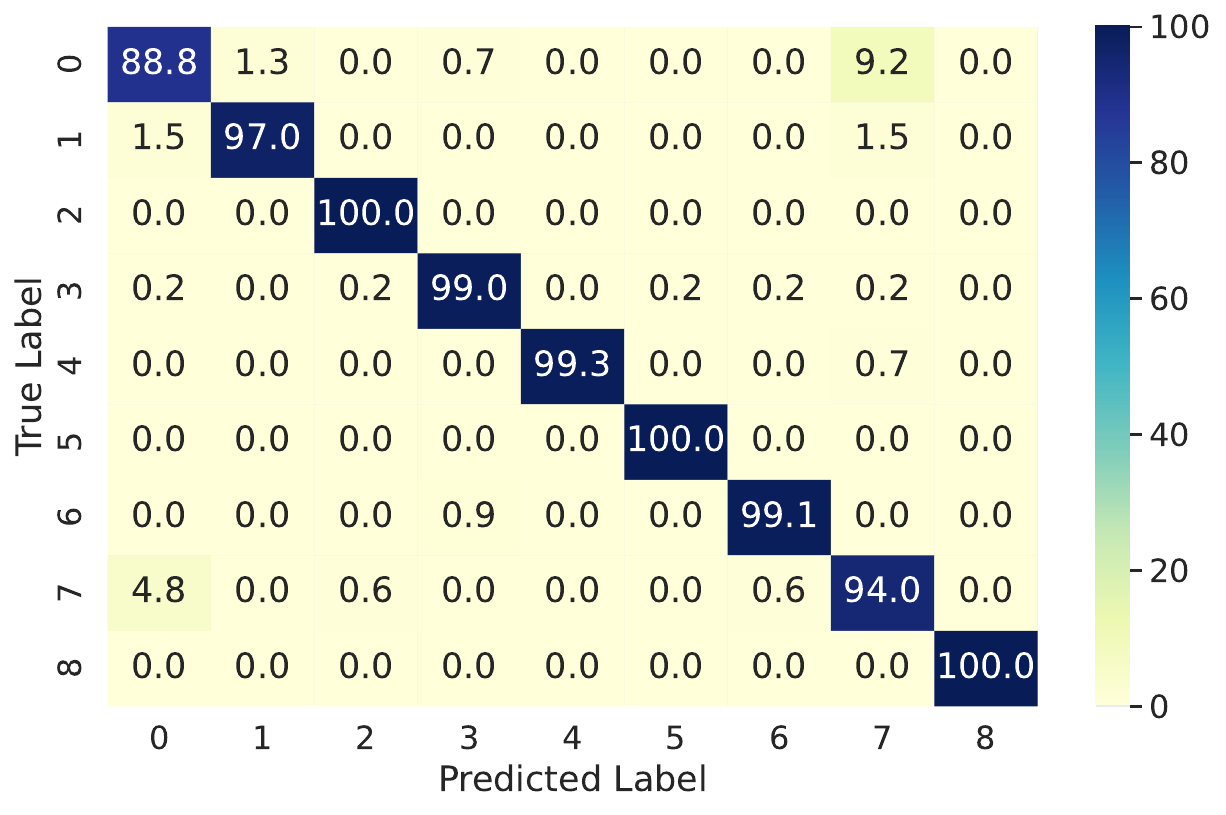}
                \caption{Optimized DA-Model}
                \label{fig:optimized-confusion-matrix}
            \end{subfigure}
            
            \caption{Comparison of classification accuracy confusion matrices: (a) Baseline model, (b) Optimized-DA model. The augmented model demonstrates improved per-class accuracies, as evident from the reduction in misclassifications illustrated in the confusion matrices. Particularly, in the case of Class 0 (Bacterial Blight) and Class 7 (Bacterial Pustule), the misclassifications have significantly reduced, highlighting the effectiveness of our approach in addressing the challenges associated with these classes.}
            \label{fig:confusion-matrices}
        \end{figure}
        
        Moreover, the iterative nature of the GA in selecting the most effective augmentations has significantly contributed to this improvement. Notably, the challenging classes of bacterial blight and bacterial pustule have shown substantial accuracy enhancements, with bacterial blight improving from 83.01\% to 88.89\%, and bacterial pustule from 85.71\% to 94.05\%. This underscores the importance of the GA's role in identifying and implementing augmentations specifically tailored to address the unique challenges posed by these classes. Overall, these results demonstrate the effectiveness of our class-specific DA approach in overcoming class-specific challenges and significantly improving classification accuracy. 
        
        In a comprehensive comparison with other automated augmentation methods on the soybean disease dataset, as detailed in Table~\ref{tab: Comparison with Other Methods}, our method notably surpasses all others in terms of accuracy. This highlights the effectiveness of our proposed approach. Importantly, our method achieves superior accuracy while significantly reducing computation requirements by only fine-tuning the baseline model for 5 epochs, without training any augmentation policy from scratch.  This streamlined approach not only enhances accuracy but also optimizes computational resources, making it a practical solution for real-world applications.

        \begin{table}[htb]
          \caption{Comparison with Other Automated Augmentation Methods}
          \label{tab: Comparison with Other Methods}
          \centering
          \begin{tabular}{l c c c}
            \toprule
            Augmentation Technique & Mean-per-class Accuracy (\%) & Sensitivity (\%) & Specificity (\%) \\
            \midrule
            AutoAugment (ImageNet) & 95.8 & 100 & 91.9 \\
            AutoAugment (CIFAR-10) & 95.5 & 97.7 & 98.5 \\
            AutoAugment (SVHN) & 95.6 & 100 & 93.9 \\
            RandAugment & 96.2 & 97.7 & 96.3 \\
            Trivial Augment & 95.9 & 99.0 & 96.9 \\
            AugMix & 95.7 & 98.4 & 97.7 \\
            \makecell[l]{GA-based Optimized DA \\ (Proposed Method)} & 97.6 & 99.2 & 97.0 \\
            \bottomrule
          \end{tabular}
        \end{table}
    
    \subsection{Impact of class-specific augmentations on misclassifications}
    
        To assess the performance of the models on misclassifications, we analyzed the confusion matrices of the baseline model and the augmented model (Figure~\ref{fig:confusion-matrices}). As mentioned earlier, the baseline model struggled particularly with predicting bacterial blight (class 0) and bacterial pustule (class 7), frequently misclassifying them interchangeably \parencite{hartman2015compendium}. However, the optimized model exhibited a noticeable reduction in misclassifications for these challenging classes. By tailoring augmentations to each class, the GA automatically selects augmentations that help distinguish these classes from each other. Consequently, the optimized model showed improved per-class accuracies, suggesting that our class-specific DA techniques effectively addressed the baseline model's limitations. These enhancements validate the effectiveness of our approach in improving classification performance, especially for the most challenging classes.

    \subsection{Comparison of optimized augmentations on different stresses}
    
\begin{figure}[t]
            \centering
            \includegraphics[width=\textwidth, trim=5 5 5 5, clip]{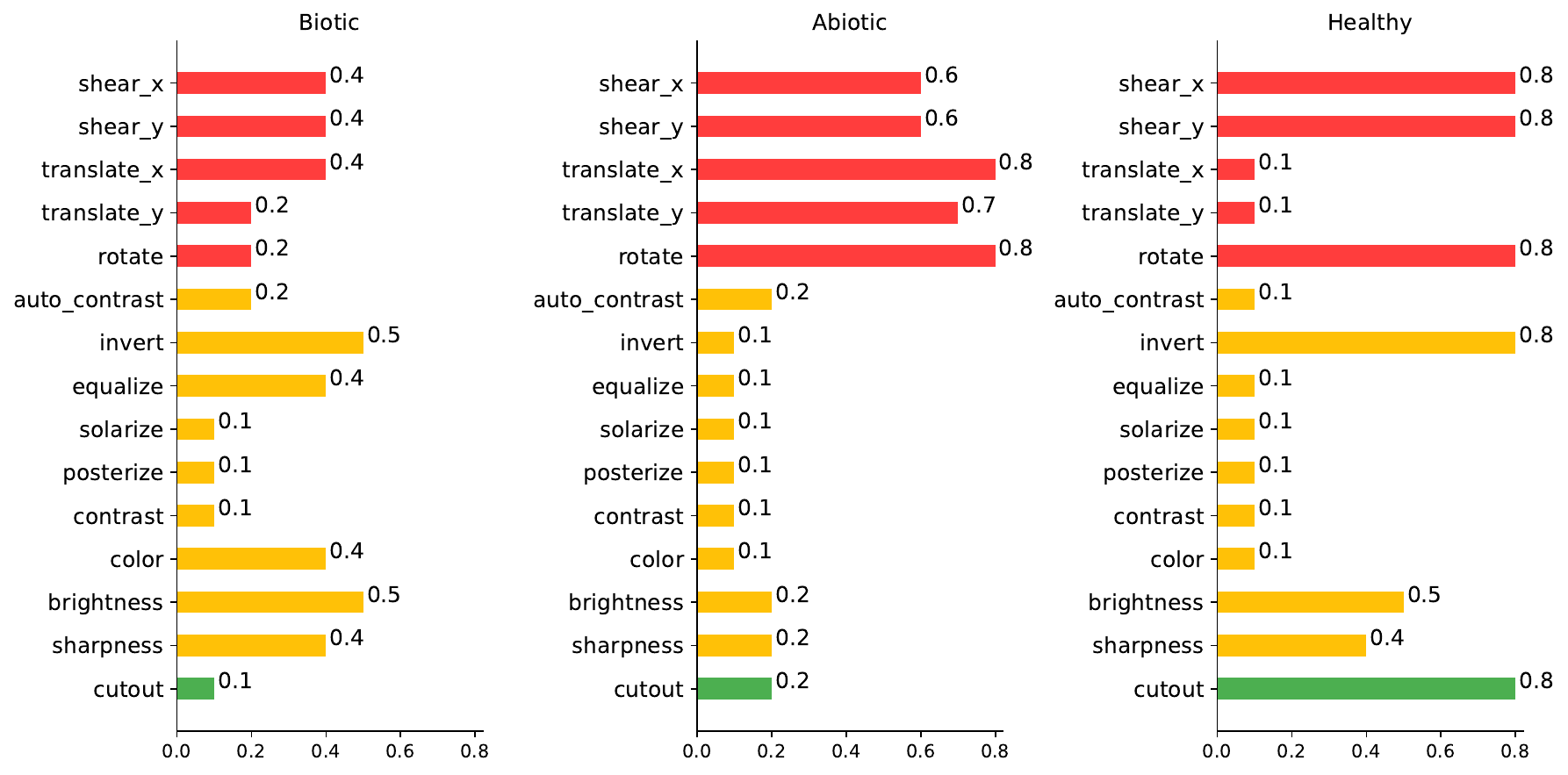}
            \caption{ Optimized augmentation policies for different stress conditions (biotic and abiotic) and healthy leaves. Augmentation techniques are categorized into three groups: Geometry-based augmentations (red), Color-based augmentations (yellow), and the Cutout (green).}
            \label{fig: augs}
        \end{figure}

        In our analysis of the optimized augmentations, we aimed to understand their impact on different stress conditions, including biotic and abiotic stresses, as well as healthy leaves. The optimized augmentation policies, depicted in Figure~\ref{fig: augs}, shows the preferences for specific augmentation types across these classes \parencite{gull2019biotic}.
        
        For biotic stress classes, we observed consideration for both color-based and geometry-based augmentations. However, specific color augmentations such as solarize, posterize, and invert were not favored due to their limited relevance to disease-related visual cues in this context. Conversely, in the case of abiotic stress classes, geometry-based augmentations were preferred, with shear, translation, and rotation being prominent choices. Additionally, augmentations like sharpness and autocontrast were selected for their effectiveness in capturing the structural changes associated with abiotic stressors.
        
        In contrast, for the healthy class, geometry-based augmentations were predominantly chosen, with brightness, sharpness, and autocontrast selected to enhance the natural appearance of healthy leaves. Interestingly, the cutout augmentation was exclusively chosen by the healthy class, while being avoided by other stress classes. This suggests that cutout augmentation, which masks specific regions in the images, may inadvertently remove relevant disease-related information for other classes. Overall, the analysis highlights the importance of selecting appropriate augmentations tailored to each stress class to improve the accuracy of soybean stress classification.

    \subsection{Does the order of augmentations matter?}
    
        To explore the impact of augmentation order on model performance, separate GA runs were conducted for each proposed augmentation sequence, categorized into three: i) Geometry-- includes augmentations that modify the geometric properties of the images, such as shearing and rotation, ii) Color-- comprises augmentations that manipulate the color and contrast characteristics of the images and, iii) Cutout. Table \ref{tab:augmentation-order} provides a summary of the results obtained from these runs. 
        
        \begin{table}[ht]
          \centering
          \caption{Accuracy Comparison of Different Order of Augmentations}
          \label{tab:augmentation-order}
          \begin{tabular}{cccc}
            \toprule
            Order of Augmentations & Accuracy (\%) \\
            \midrule
            Geometry $\to$ Color $\to$ Cutout & 97.6 \\
            Geometry $\to$ Cutout $\to$ Color & 96.7 \\
            Color $\to$ Geometry $\to$ Cutout & 96.5 \\
            Cutout $\to$ Geometry $\to$ Color & 96.4 \\
            Color $\to$ Cutout $\to$ Geometry & 96.4 \\
            Cutout $\to$ Color $\to$ Geometry & 95.9 \\
            \bottomrule
          \end{tabular}
        \end{table}
        
        We observed that the augmentation order has a slight influence on the model's performance. The highest accuracy of 97.6\% was achieved when the augmentations were applied in the order of Geometry, followed by Color, and then Cutout \parencite{perez2017effectiveness}. This implies that initiating the augmentation process with geometric transformations, followed by color manipulations, and concluding with cutout techniques can lead to superior accuracy in the context of our dataset. It's important to note that these findings are specific to our dataset, and results may vary for different datasets. Despite small relative change in accuracies, the overall performance remained consistently high across all orders, indicating that the choice of augmentation order may not be critical in achieving strong results, and further research is needed in other datasets. These findings suggest that while the order of augmentations may have a marginal impact on the model's performance, the selection and combination of augmentation techniques play a more significant role in improving accuracy.

\section{ Discussion}

    DL models often struggle to achieve consistent high performance across all classes within a dataset, despite achieving high overall accuracy. Data imbalances and lack of diversity in the training data are among the key reasons for this phenomenon. Data augmentation, which aim to enhance model performance and mitigate the challenges imposed by data imbalances and diversity, have emerged as an effective approach to address these issues. In this study, we demonstrate that by tailoring augmentations specific to each class in a dataset, these limitations can be effectively mitigated.
    
    To achieve this, we deployed tailored augmentations for each class in our soybean disease dataset using GA-based optimization. We fine-tuned a well-trained baseline model for each data augmentation policy generated by the GA. Through comprehensive evaluation of mean per-class accuracy and confusion matrices, we observed significant improvements in the accuracy of each class in the dataset. Notably, the accuracy of confounding classes, such as bacterial blight and bacterial pustule, has also been substantially improved.
    
    The key mechanism behind our approach lies in the utilization of GA to automatically select augmentations tailored to each class. The GA iteratively explores the augmentation space and identifies the most effective augmentations that maximize the mean per-class accuracy. By fine-tuning the baseline model with these augmentations for a limited number of epochs, we efficiently enhance the model's ability to distinguish between different classes and improve overall classification performance. This adaptive and iterative approach bears resemblance to boosting techniques in ML \parencite{tanha2020boosting}. Just as boosting algorithms iteratively train weak learners to create a strong ensemble model that excels in classifying difficult instances, our method iteratively refines the baseline model by selecting augmentations tailored to address the challenges posed by specific classes. The GA's exploration of the augmentation space parallels the boosting process of focusing on misclassified instances in successive iterations, ultimately leading to improved classification performance.

    Moreover, our approach offers two distinct advantages over existing techniques \parencite{cubuk2018autoaugment, ho2019population, lim2019fast, marrie2023slack, }. Firstly, unlike previous methods that optimize augmentations for the entire dataset, we tailor augmentations specific to each class in the dataset. This class-specific nature allows our approach to address the unique characteristics and challenges associated with individual classes, resulting in improved performance across all classes. Secondly, our method significantly reduces computation requirements by only fine-tuning the last layer of the model for a limited number of epochs for each augmentation policy generated by the GA. This streamlined approach not only enhances accuracy but also optimizes computational resources, making it a practical solution for real-world applications
    
    Furthermore, our analysis of the augmentations selected by the optimized model reveals interesting insights into the preferences of specific stresses (biotic, abiotic, and healthy) for particular augmentation types. This verifies that classes within a dataset can indeed have different preferences for augmentations, highlighting the importance of class-specific augmentation strategies.
    
    Additionally, we investigated the effect of the order in which augmentations are applied on model performance. Our results indicate that while the augmentation order may have a slight influence on performance, the selection and combination of augmentation techniques play a more significant role in improving accuracy. Initiating the augmentation process with geometric transformations, followed by color manipulations, yielded superior accuracy in our dataset.
    
    Overall, our study underscores the effectiveness of tailored class-specific data augmentations in enhancing DL model performance for soybean stress classification. By addressing class-specific challenges and optimizing the augmentation process, our approach offers a promising solution for accurate disease diagnosis and management in agricultural applications. Future research directions may involve exploring the application of our method to different crops and stress conditions, as well as investigating the integration of advanced ML techniques for further performance enhancement.

\section{Conclusion}

    This study demonstrates the efficacy of a GA-based approach in identifying class-specific augmentations to improve plant stress classification accuracy. By fine-tuning a baseline model with tailored augmentations, we achieved a notable increase in mean-per-class accuracy, with the optimized model achieving an impressive average per-class accuracy of 97.61\%, surpassing the performance of existing automated augmentation methods. Particularly, previously challenging classes such as bacterial blight and bacterial pustule showed significant accuracy enhancements, with bacterial blight accuracy increasing from 83.01\% to 88.89\% and bacterial pustule accuracy jumping from 85.71\% to 94.05\%. These improvements highlight the effectiveness of our approach in addressing class-specific challenges.

    The findings of this study underscore the importance of tailored augmentation strategies for individual classes in plant stress classification tasks. Leveraging GA optimization, we showcased significant improvements in accuracy, providing valuable insights for the development of class-specific augmentation techniques. These results have implications beyond soybean disease classification, offering guidance for similar classification tasks in agriculture and other domains.

\section*{Acknowledgements}

    This work was supported by the AI Institute for Resilient Agriculture (USDA-NIFA 2021-67021-35329) and COALESCE: COntext Aware LEarning for Sustainable CybEr-Agricultural Systems (NSF CPS Frontier 1954556, FACT: A Scalable Cyber Ecosystem for Acquisition, Curation, and Analysis of Multispectral UAV Image Data (USDA-NIFA \#2019-67021-29938), Smart Integrated Farm Network for Rural Agricultural Communities (SIRAC) (NSF S\&CC \#1952045), Iowa Soybean Association, Plant Sciences Institute, R F Baker Center for Plant Breeding, and USDA CRIS Project IOW04717.

\section*{Author Contributions}
    N.S. and B.G. conceived the project. N.S. ran the experiments. N.S., A.B., and Z.J. conducted the augmentation analysis. N.S. and A. B. wrote the paper. All authors reviewed and edited the manuscript. 

\section*{Competing Interests}
    The authors declare that there is no conflict of interest regarding the publication of this article.
\section*{Data Availability}
    The data used for the creation of this manuscript is available at \textit{https://github.com/nasla-96/Class-specific-Data-Augmentation-for-Plant-Stress-Classification/tree/master}

\section*{Supplemental files}
\begin{itemize}
    \item Supplemental table S1: Data Summary - Number of samples in the training, validation, and test sets for each class.
    \item Supplemental figure S1: Principal Component Analysis (PCA) Plot after Histogram of Oriented Gradients (HOG) feature extraction for each class from class 0 to class 8.
    \item Supplemental table S2: The performance of different CNN architectures on the soybean leaf stress dataset.
    \item Supplemental table S3: 5-fold cross-validation of the class-specific DA model.
\end{itemize}

\printbibliography

\end{document}


\begin{center}
{\usefont{OT1}{phv}{b}{}\selectfont\Large{Supplementary File}}

\vspace{0.1in}

{\usefont{OT1}{phv}{}{}\selectfont\normalsize
{Class-specific Data Augmentation for Plant Stress Classification
}}
\end{center}

Table S1 provides a summary of the dataset used in the study, displaying the number of
180 images allocated to the training, validation, and test datasets for each class. The training, validation, and test datasets were composed of 13, 420 (80\%), 1, 491 (9\%), and 1, 662 (11\%), 179 respectively.

\begin{table}[h]
  \centering
  \caption*{Supplemental table S1: Data Summary - Number of samples in the training, validation, and test sets for each class.}
  \begin{tabular}{cccc}
    \toprule
    Class & Train & Valid & Test \\
    \midrule
    0 & 1234 & 137 & 153 \\
    1 & 1100 & 122 & 136 \\
    2 & 908 & 101 & 113 \\
    3 & 3420 & 380 & 423 \\
    4 & 1130 & 125 & 140 \\
    5 & 1493 & 166 & 185 \\
    6 & 1770 & 197 & 219 \\
    7 & 1355 & 151 & 168 \\
    8 & 1010 & 112 & 125 \\
    \midrule
    Total & 13420 (81\%) & 1491 (9\%) & 1662 (10\%) \\
    \bottomrule
  \end{tabular}
  \label{tab:data}
\end{table}

We conducted an analysis of image diversity across the training, validation, and testing sets of our data using Histogram of Oriented Gradients (HOG) features and Principal Component Analysis (PCA). HOG features were extracted from images in the training, validation, and test sets for each class, capturing edge and gradient structures robustly. Subsequently, PCA was applied to reduce the high-dimensional HOG features to two principal components, enabling visualization and comparison of dataset diversity. The PCA scatter plots, as depicted in Figure S1, revealed that images from all sets were interspersed across the feature space, indicating similar variability coverage across sets and no isolated clustering. This overlap suggests diverse and representative datasets, mitigating sampling biases and ensuring reliable model performance metrics across cross-validation folds. 

\begin{figure}[ht]
   \includegraphics[width=0.9\textwidth, trim=10 215 12 110, clip]{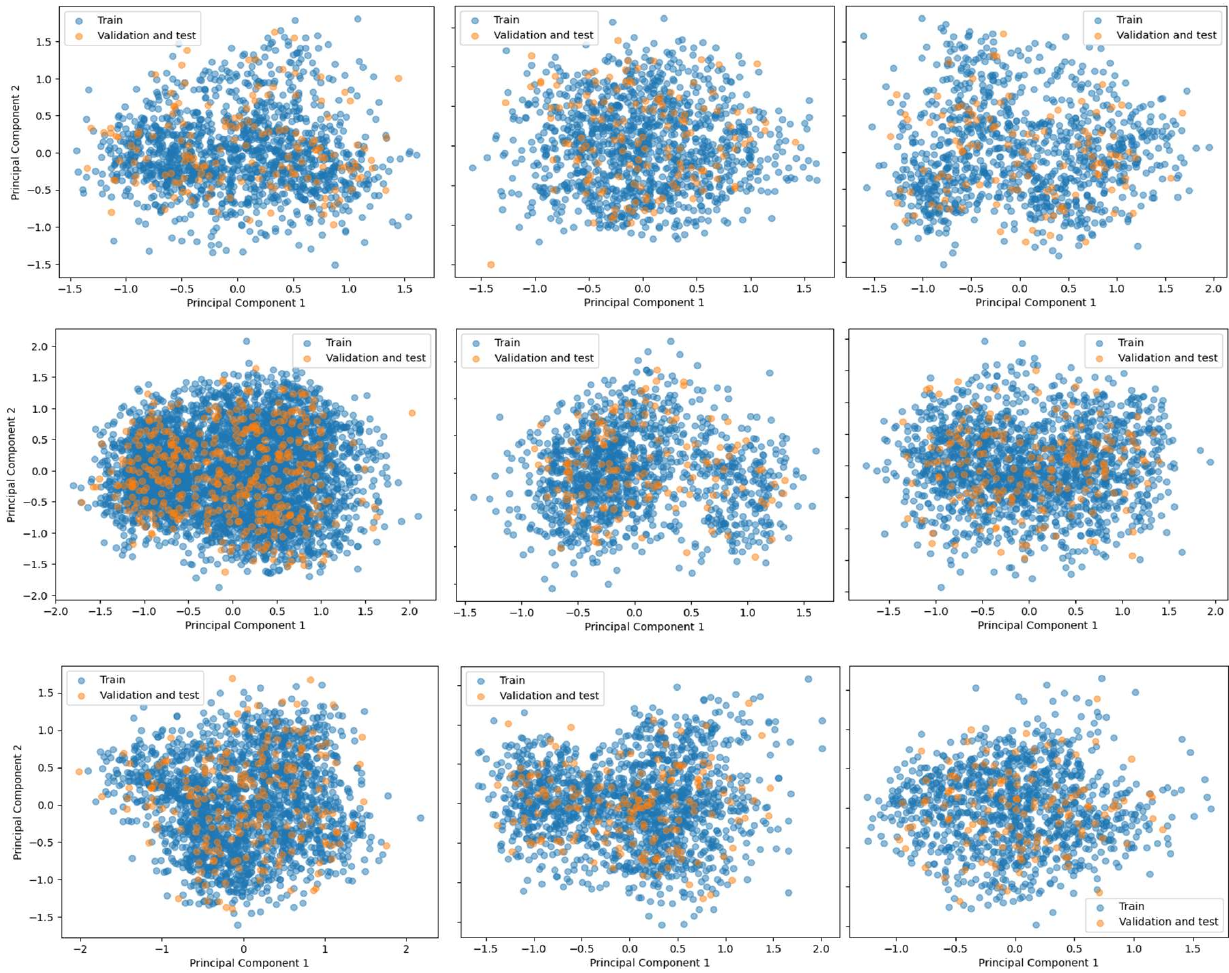}
    \caption*{Supplemental figure S1: Principal Component Analysis (PCA) Plot after Histogram of Oriented Gradients (HOG) Feature Extraction for each class from class 0 to class 8.
        \label{fig: data diversity}}
\end{figure}

To establish a fair and comprehensive baseline for soybean stress classification, we followed the methodology outlined in (Ghosal et al. 2018), utilizing the same dataset and baseline model. However, recognizing the potential for further improvement, we sought to explore different deep learning architectures to enhance the baseline accuracy. The results of our investigation, as shown in table S2, indicated that ResNet50 achieved the highest accuracy among the tested architectures, prompting its selection for further experimentation and evaluation. The baseline model was trained for 350 epochs on the dataset
 
\begin{table}[h]
  \centering
  \caption*{Supplemental table S2: The performance of different CNN architectures on the soybean leaf stress dataset}
  \label{tab:accuracies}
  \begin{tabular}{cccc}
    \toprule
    Model & Mean-per-class Accuracy (\%) & Sensitivity (\%) & Specificity (\%)\\
    \midrule
    DCNN & 93.42 & 94.0 & 96.6\\
    VGG16 & 94.60 & 90.6 & 99.2\\
    VGG19 & 94.62 & 98.5 & 93.5\\
    ResNet18 & 94.85 & 100 & 91.4\\
    ResNet34 & 94.92 & 100 & 94.1\\
    ResNet50 & 95.09 & 97.7 & 96.8\\
    \bottomrule
  \end{tabular}
\end{table}

Table S3 provides the results of a 5-fold cross-validation conducted to assess the stability and performance of the class-specific DA) model. The cross-validation was performed to evaluate the model's ability to accurately classify instances into predefined classes while ensuring robustness and generalizability.

\begin{table}[h]
  \caption*{Supplemental table S3: 5-fold cross validation of the class-specific DA model}
  \label{tab: cross validation}
  \centering
  \begin{tabular}{cccc}
    \toprule
    Fold Number & Mean-per-class Accuracy (\%) & Sensitivity (\%) & Specificity (\%) \\
    \midrule
    1 & 97.0 & 97.0 & 98.5 \\
    2 & 97.6 & 99.2 & 97.0 \\
    3 & 97.3 & 100 & 96.2 \\
    4 & 96.4 & 99.2 & 96.9 \\
    5 & 97.6 & 98.5 & 98.5 \\
    \bottomrule
  \end{tabular}
\end{table}